\documentclass[lettersize,journal]{IEEEtran}
\usepackage{amsmath,amsfonts,amssymb}
\usepackage{algorithmicx}
\usepackage{algorithm}
\usepackage{array}
\usepackage[caption=false,font=normalsize,labelfont=sf,textfont=sf]{subfig}
\usepackage{textcomp}
\usepackage{stfloats}
\usepackage{url}
\usepackage{verbatim}
\usepackage{graphicx}
\usepackage{cite}
\usepackage{makecell}
\usepackage{algpseudocode}
\usepackage[dvipsnames]{xcolor}
\usepackage{tabularx}
\usepackage{multirow}

\usepackage[hidelinks]{hyperref}
\hypersetup{
    colorlinks,
    linkcolor={red},
    citecolor={blue},
    urlcolor={blue}
}

\usepackage{booktabs}
\usepackage{caption}
\newcommand{\tildecorrect}{\raisebox{0.5ex}{\texttildelow}}
\newcommand{\emitdatasetabb}{EMIT-MSeg}

\begin{document}

\title{A Fast Methane Detection Pipeline on Board Satellites Based on Mag1c-SAS and LinkNet}

\author{Jonáš Herec, Vít Růžička, Rado Pitoňák, Jan Sedmidubsky
\thanks{J. Herec and R. Pitoňák are with Zaitra s.r.o., Brno, Czech Republic.}
\thanks{V. Růžička is with NASA JPL, California, USA.}
\thanks{J. Herec and J. Sedmidubsky are with the Faculty of Informatics, Masaryk University, Brno, Czech Republic.}}

\maketitle

\begin{abstract}
Methane is a potent greenhouse gas, and detecting its leaks early via hyperspectral satellite imagery can help climate change mitigation efforts. Meanwhile, many existing hyperspectral missions only capture areas manually targeted by operators, thus missing potential events of interest. To overcome slow downlink rates cost-effectively, onboard detection is a viable solution. However, traditional methane detection methods are too computationally demanding for resource-limited onboard hardware. This work accelerates methane detection by focusing on efficient, low-power algorithms. In particular, we test fast target detection ACE and CEM methods that have not been previously used for methane detection and propose Mag1c-SAS -- a significantly faster variant of the current state-of-the-art Mag1c algorithm for methane detection. To explore their true detection potential, we integrate them with a machine learning model based on U-Net and LinkNet. We evaluate our methods on the STARCOP dataset and a novel \emitdatasetabb{} dataset, which we introduce and open-source in this study alongside a high-quality annotation strategy. The proposed Mag1c-SAS approach proves highly effective by operating \tildecorrect80$\times$ faster than the original Mag1c approach, providing a visually similar, but noisier result. When additionally paired with the lightweight LinkNet approach, it effectively reduces noise, achieving AUPRC score improvements of over 30\,pp on \emitdatasetabb{} compared to the baseline Mag1c approach, and an F1 score on STARCOP \tildecorrect4\,pp higher. We also evaluate two novel band selection strategies and confirm the system's onboard viability through hardware profiling, demonstrating marginal power consumption and efficient CPU/RAM utilization. We release the final system in a user-friendly and lightweight PyPI library at: \url{https://pypi.org/project/onboard-methane-detection/}, alongside all experimental code, models, and data at: \url{https://github.com/zaitra/methane-filters-benchmark}.
\end{abstract}


\section{Introduction}
\IEEEPARstart{M}{ethane}  is the second most significant greenhouse gas contributing to global warming after $\text{CO}_2$~\cite{kuylenstierna2021global_UN}. While $\text{CO}_2$ is more prevalent, methane is more potent, mainly in the short term (about 84--87 times more than $\text{CO}_2$ over a 20-year period~\cite{Intergovernmental}). A significant share of methane emissions in the oil and gas industry originates from episodic ultra-emission events, often caused by equipment failures and other unpredictable factors~\cite{lauvaux2022global}.

These events could be detected and mitigated early with the use of satellite monitoring, but it is a challenging task. While methane has a distinct spectral signature with its strongest absorption features in the \tildecorrect2100--2500 nm range, the change in measured radiance is too subtle and cannot be detected by the human eye. However, it can be seen in data from current imaging spectroscopy sensors such as NASA's AVIRIS-NG or EMIT~\cite{green2022_EMIT} and from the sensors of planned missions such as NASA's upcoming EAGLE-VSWIR~\cite{cawse2021nasa_SBG}, or the ESA's CHIME~\cite{rast2021copernicus_CHIME}. These missions carry hyperspectral instruments that nearly continuously capture light across hundreds of spectral channels, allowing methane's spectral signature to be leveraged for its detection.

A large number of channels is advantageous, but it also hinders timely data delivery. Downlinking from satellites is often slow, as data is transmitted via limited-bandwidth radio communication only a few times per day when the satellite passes over a ground station. Increasing transmission speed would significantly raise costs, not only due to the added weight and expense of more powerful satellite antennas, but also because it would require a larger network of ground stations.
The approach of many currently flown satellites (for example, PRISMA \cite{cogliati2021prisma}, EnMAP \cite{guanter2015enmap} or Tanager-1 \cite{duren2025carbon_tanager}) is to observe only over manually selected regions, which severely reduces the number of potentially captured methane leak events.

To address this, algorithms and machine learning (ML) models can be employed on board, as already explored for disaster events in \cite{RaVAEn, herec2026sttorm}, and specifically for methane detection in \cite{Ruzicka2025HyperspectralViTs}. However, since the goal is to reduce the cost and improve the timely delivery of salient data, these models must be designed to fit the constraints of the mission without compromising its objectives. Given the high power requirements of GPUs and their ongoing adaptation for the space environment, the key could be developing efficient algorithms that can run on commonly used solutions~\cite{nasasmallsats} -- lower-power CPUs or FPGAs~\cite{guerrieri2018fpgas_for_space}.

The key part of the pipeline is the computation of the methane enhancement product, which generates a 2D image representing the detection of the methane spectral signature from the 3D hyperspectral cube. This process effectively reduces the number of channels from dozens/hundreds to just one, which can then be used for real-time methane detection by thresholding or by an efficient machine learning model. While not previously computed on board satellites, it has been shown that calculating this product in real time on aircraft helped successfully guide ground teams toward particularly strong leaks \cite{mf2}.

\begin{figure}[ht!]
\centerline{\includegraphics[width=0.5\textwidth]{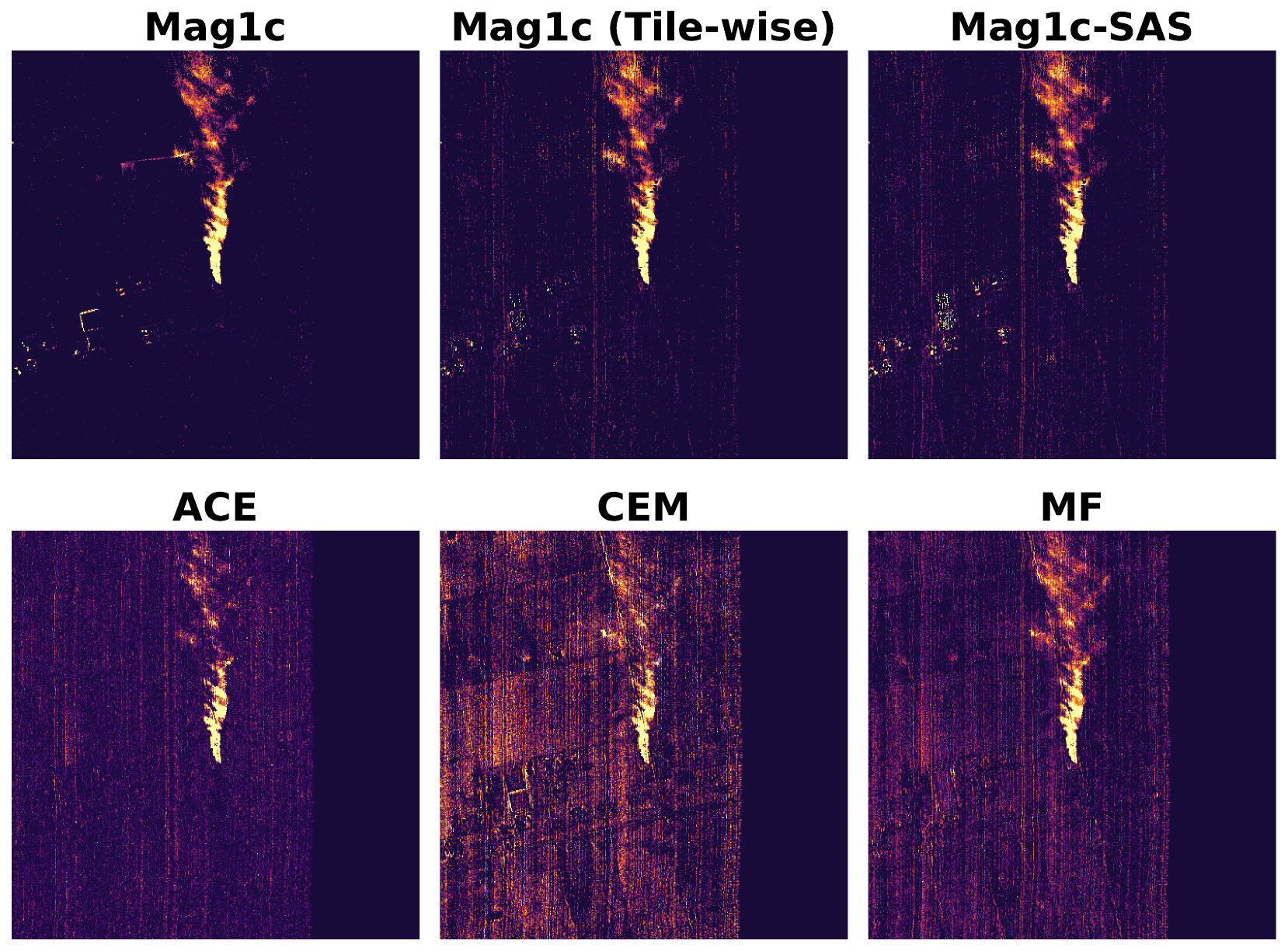}}
\caption{Visualization of a strong methane plume event showing all methane enhancement products used in this study. Note that visual scaling was adjusted manually to aid legibility.}
\label{direct}
\vspace{-6mm}
\end{figure}

This is typically solved by the Matched Filter (MF)~\cite{tiemann2024machine} product, with the Matched filter with Albedo correction and reweiGhted L1 sparsity Code (Mag1c)~\cite{mag1c} being a state-of-the-art variant. These products are calculated for each column of the input image; in Mag1c, this is further repeated in several iterations. We argue that these products are prohibitively slow for onboard deployment. Therefore, we explore faster and simpler variants of these methane enhancement products, where, besides other optimizations, we compute the whole filter at once instead of column-wise, to further speed up the whole process.

Since the number of bands is a crucial parameter for these algorithms, we experiment with newly proposed band selection strategies to determine if the runtime can be further reduced by selecting fewer bands without compromising accuracy. We trade the quality of the products for speed, but this can be mitigated by lightweight machine learning models, which refine the methane product, similarly to \cite{jongaramrungruang_methanet_2022, ruuvzivcka2023semantic}. Also, we focus on more easily detectable large plumes, which are particularly relevant, as their impact is disproportionately larger than that of smaller leaks \cite{duren_californias_2019} and is thus more critical for onboard detection. To evaluate these optimizations and validate our models across diverse conditions, we introduce a new testing dataset (called \emitdatasetabb{}), developed with a high-quality annotation strategy.
Specifically, we build on our preliminary work~\cite{herec2025edhpc} and significantly extend it, which overall results in the following contributions.
\begin{itemize}
    \item We propose a new methane product called ``Mag1c Sped up with Additional Sparsity'' (\textbf{Mag1c-SAS}), which is \tildecorrect80$\times$ faster than the original Mag1c applied to columns, with an acceptable drop in accuracy.
    \item We benchmark two newly proposed band selection strategies (Highest Transmittance and Variance Increase) and test the influence of the number of channels on accuracy and speed.
    \item We investigate various target detection methods, including the Constrained Energy Minimization (CEM) method \cite{harsanyi1993detection}, which is \tildecorrect140\,$\times$ faster than the original Mag1c approach applied to columns.
    \item We train a segmentation ML model to refine these products, consistently improving the F1 score by up to \tildecorrect20\,pp. For training, we use the STARCOP dataset derived from airborne AVIRIS-NG data.
    \item Alongside a high-quality annotation strategy, we introduce a new methane-segmentation dataset (\textbf{\emitdatasetabb{}}) derived from spaceborne EMIT data. This allows us to evaluate model performance on an unseen sensor.
    \item We release this framework as an easy-to-use PyPI library. To our knowledge, it is the first sensor-agnostic methane detection pipeline for low-power satellite CPUs, achieved by combining an accelerated matched filter with an ML model. We also provide extensive hardware profiling of the library on a space-grade Xiphos Q8J.
\end{itemize}

\section{Methodology}

\subsection{Methane enhancement products}
\label{products}

Figure \ref{direct} shows a selection of methane enhancement products explored in this work. 
A widely used method for methane detection is the matched filter~\cite{manolakis2003hyperspectral}, often implemented with various modifications. Its primary goal is to identify and amplify the methane signal while suppressing background noise. This is accomplished by comparing the spectral signature to deviations from the band-wise means at each hyperspectral pixel, while considering the background characteristics modeled by the band's variance-covariance matrix.

Here, we use the \textbf{matched filter (MF)} as follows:
\[
y_i = \frac{(x_i - \mu)^\top \, C^{-1} (t \cdot \mu)}{(t \cdot \mu)^\top C^{-1} (t \cdot \mu)}
\]
\noindent where:
\begin{itemize}
  \item \( x_i \in \mathbb{R}^{p} \) is the hyperspectral pixel with dimensionality \( p \) (number of bands);
  \item \( t \in \mathbb{R}^{p} \) is the target spectrum to detect;
  \item \( \mu \in \mathbb{R}^{p} \) is the vector of band-wise means, i.e.,
\[
\mu = \frac{1}{N} \sum_{i=1}^{N} x_i;
\]
  \item \( C \in \mathbb{R}^{p \times p} \) is the covariance matrix of the mean-centered data:
\[
C = \frac{1}{N - 1} \sum_{i=1}^{N} (x_i - \mu)(x_i - \mu)^\top;
\]
  \item \(y_i\) is the matched filter response at a pixel: a scalar, indicating how well this pixel matches the target spectrum.
\end{itemize}

This definition is the same as in \cite{matchedfilteroriginal}, with one adjustment that normalizes the target spectrum using a multiplicative model ($t \cdot \mu$) rather than an additive one ($t - \mu$) as in \cite{matchedfilteroriginal}. This change is necessary because traditional target detection methods have been designed for solid, opaque materials like minerals, where the target's signal mixes additively with the background. In contrast, detecting a transparent gas like methane relies on a multiplicative relationship; the gas does not sit on top of the background, but absorbs the background radiation passing through it, as described by the Beer-Lambert law~\cite{mag1c}.

Very similar to the matched filter is the \textbf{Constrained Energy Minimization (CEM)} detector~\cite{harsanyi1993detection}. While it originates from a different theoretical background than MF, it arrives at an almost identical filter definition, but without the whitening (mean subtraction) of the data. The CEM detector is defined as:
\[
y_i = \frac{x_i^\top \, K^{-1} t}{t^\top K^{-1} t}
\]
where $K$ (often referred to as sample correlation matrix or auto-correlation matrix) is similar to $C$, but it is computed directly from the raw (non-centered) data as:
\[
K = \frac{1}{N - 1} \sum_{i=1}^{N} x_i x_i^\top.
\]
The last classic target detection method used in this study is the \textbf{Adaptive Cosine Estimator (ACE)}~\cite{10.1117/12.818790}. Here, we also apply multiplicative rather than additive target normalization, resulting in this definition:
\[
y_i = \frac{\left[(x_i - \mu)^\top C^{-1} (t \cdot \mu)\right]^2}{\left[(t \cdot \mu)^\top C^{-1} (t \cdot \mu)\right] \cdot \left[(x_i - \mu)^\top C^{-1} (x_i - \mu)\right]}
\]
Again, its core principles are very similar to MF and CEM, but it has additional normalization by the value of the pixel itself. Thus, two pixels (vectors) with the same spectral response (direction) but different brightness (magnitude) will receive the same score.
 
While these methods are typically used as baselines in target detection research, they could be ideal candidates for this application. This is because more advanced target detection methods often offer only a slight increase in accuracy but result in a significant increase in processing time~\cite{slightaccuracyformorespeed}.

For remote sensing imagery, the matched filter was particularly adapted. In these applications~\cite{mf1,mf2}, the algorithm is applied to each column separately, so the key parameters (\(\mu, C\)) are estimated separately for each sensor in a push-broom instrument. Due to calibration and manufacturing differences of the sensors, this approach ensures more robust and accurate estimation of the background characteristics, improving the accuracy and eliminating artifacts like striping in the final output.

However, this involves the calculation of band means, the construction and inversion of the variance-covariance matrix for each column, which can be highly computationally demanding. Additionally, one of the most widely used and accurate methods, Mag1c~\cite{mag1c}, further increases computational complexity by performing an iterative calculation of the filter. There, the means and variance-covariance matrix are iteratively re-estimated from data after removing the methane enhancement from the previous iteration, as estimating them from methane-contaminated data inhibits the filter's detection capabilities. 

The original Mag1c pseudocode is shown in Algorithm~\ref{alg:spatialrwl1mf}. Its input arguments are: the hyperspectral datacube $D \in \mathbb{R}^{H \times W \times p}$, where $H$ is an image height, $W$ is an image width, and $p$ is the number of bands; a methane spectrum $s \in \mathbb{R}^{p}$; and the number of iterations $N_{\mathrm{iter}}$. Detection, however, is performed independently for each column $L \in \mathbb{R}^{H \times p}$. The core in row \ref{core-principle} can be expanded as:
\[
\alpha_i^{k} = \max \left( 
\frac{
\left( \boldsymbol{L}_i - \boldsymbol{\mu}^k \right)^T 
{\boldsymbol{C}^k}^{-1} 
\left( \boldsymbol{\mu}^{k} \odot \boldsymbol{s} \right) 
- w_i^k
}{
r_i 
\left( 
\left( \boldsymbol{\mu}^{k} \odot \boldsymbol{s} \right)^T 
{\boldsymbol{C}^k}^{-1} 
\left( \boldsymbol{\mu}^{k} \odot \boldsymbol{s} \right) 
\right) 
},\ 0 
\right).
\]

If we set the sparsity-enforcing term $w_i^k = 0$ and the Albedo correction term $r_i = 1$, this formulation becomes equivalent to the previously described matched filter followed by a ReLU activation function.

However, even though the core principle is similar, the algorithm results differ mainly due to the following three changes:
\begin{enumerate}
\item It uses the Albedo correction term $r_i = \frac{
                                        \boldsymbol{L}_i^T \boldsymbol{\mu}
                                    }{
                                        \boldsymbol{\mu}^T \boldsymbol{\mu}
                                    }$, which is used in the product denominator. Methane detections, which are stronger due to the greater reflectivity of their background (brighter pixels), are thus normalized by a greater value.
\item It uses a sparsity-enforcing term $\boldsymbol{w}^k = \frac{1}{\textcolor{red}{r} (\boldsymbol{\alpha}^{k-1} + \epsilon)}$, which is subtracted from the methane detection in the numerator. Combined with the ReLU function and the iterative nature of Mag1c, this term pushes small and likely false-positive detections to zero.
\item As the background characteristics (\(\mu, C\)) should be methane free to enable correct methane signal amplification, the filter is computed in $N_{\mathrm{iter}}$ iterations, where in each iteration, the previous methane detection is subtracted from the pixel during the parameters computation, as seen in rows \ref{first-row}--\ref{last-row}.
\end{enumerate}
Besides the original, we explore two accelerated variants of the Mag1c product:

\begin{itemize}
\item \textbf{Mag1c (tile-wise):} Instead of applying Mag1c column-wise, we apply it to the whole tile at once to speed up the computation on the low-power CPU.
\item \textbf{Mag1c-SAS:} The computation is also performed over the entire tile, but split into two stages. The first stage involves computationally heavy iterative parameter estimation, but this is run on only a small fraction ($f = 1\,\%$) of the tile. In the second stage, the Mag1c-SAS product is computed across the full tile using the parameters estimated in the first stage. Although iterative computation is still necessary in the second stage to enforce sparsity, the matched filter is computed only once, as each subsequent iteration only recalculates, normalizes, and subtracts the sparsity-enforcing term ($\boldsymbol{w}^k$) from the filter output. These changes yield a slightly less precise but substantially faster algorithm, making it well-suited for low-power CPUs. The full customized algorithm is detailed in Algorithm~\ref{alg:mag1c-sas}. The $f$-subset is selected using uniformly spaced steps over the flattened tile to ensure both representativeness and reproducibility. The value $f = 1\,\%$ was intentionally selected to achieve a $100\times$ reduction in the dominant $O(N \cdot p^2)$ variance-covariance matrix computation.
\end{itemize}

\begin{figure}[ht!]
    \vspace{-5mm}
    \centering
    \begin{algorithm}[H]
        \caption{Original Mag1c algorithm. Content in \textcolor{red}{red} was added or modified for clarity, reflecting implementation details present in the code but omitted from the original paper (see Figure 1 in \cite{mag1c}).}
        \label{alg:spatialrwl1mf}
        \begin{algorithmic}[1]
        
            {\color{red}
            \Procedure{Mag1c}{$\boldsymbol{D}$, $\boldsymbol{s}$, $N_{\mathrm{iter}}$}
                \State Initialize $\boldsymbol{A}$ as $H \times W$ matrix
                \For{$j = 1$ \textbf{to} $W$}
                    \State Extract column $j$ from $\boldsymbol{D}$: $\boldsymbol{d}_j \in \mathbb{R}^{H \times p}$
                    \State $\boldsymbol{\alpha}_j \gets \textsc{AlbedoReWeightL1Filter}(\boldsymbol{d}_j, \boldsymbol{s}, N_{\mathrm{iter}})$
                    \State $\boldsymbol{A}[:, j] \gets \boldsymbol{\alpha}_j$
                \EndFor
                \State \textbf{return} $\boldsymbol{A}$
            \EndProcedure}
            \Procedure{AlbedoReWeightL1Filter}{\textcolor{red}{$\boldsymbol{L} $}, $\boldsymbol{s}$, $N_{\mathrm{iter}}$}
            \State Initialize $\boldsymbol{\mu}^0 = \frac{1}{N} \sum_{i}^{N} \boldsymbol{L}_i$
            \State Initialize $\boldsymbol{C}^0 =
                                \frac{1}{N} \sum_{i}^{N}
                                \left(
                                    \boldsymbol{L}_i - \boldsymbol{\mu}^0
                                \right)
                                \left(
                                    \boldsymbol{L}_i   - \boldsymbol{\mu}^0
                                \right)^T$
            \For{$i = 1$ \textbf{to} $N$}
                \State Set $r_i = \frac{
                                        \boldsymbol{L}_i^T \boldsymbol{\mu}
                                    }{
                                        \boldsymbol{\mu}^T \boldsymbol{\mu}
                                    }$
                \State Initialize $ \alpha^0_i =
                                        \frac{
                                            \left(
                                                \boldsymbol{L}_i - \boldsymbol{\mu}^0
                                            \right)^T
                                            {\boldsymbol{C}^0}^{-1}
                                            \left(
                                                \boldsymbol{\mu}^0 \odot \boldsymbol{s}
                                            \right)
                                        }{
                                            r_i
                                            \left(
                                                \boldsymbol{\mu}^0 \odot \boldsymbol{s}
                                            \right)^T
                                            {\boldsymbol{C}^0}^{-1}
                                            \left(
                                                \boldsymbol{\mu}^0 \odot \boldsymbol{s}
                                            \right)
                                        } $
            \EndFor
            \For{$k = 1$ \textbf{to} $N_{\mathrm{iter}}$} \color{red}
                \State \color{black} $\boldsymbol{w}^k = \frac{1}{\textcolor{red}{r} (\boldsymbol{\alpha}^{k-1} + \epsilon)}$
                \State $\boldsymbol{\mu}^k = 
                            \frac{1}{N} \sum_{i}^{N} 
                                \left(
                                    \boldsymbol{L}_i - r_i \alpha^{k-1}_i \boldsymbol{\mu}^{k-1} \odot \boldsymbol{s}   
                                \right)$ \label{first-row}
                \For{$i = 1$ \textbf{to} $N$}
                    \State Let $\boldsymbol{d}_{Ci} = 
                        \boldsymbol{L}_i 
                        - r_i \alpha^{k-1}_i \boldsymbol{\mu}^{k} \odot \boldsymbol{s}  
                        - \boldsymbol{\mu}^k$
                \EndFor
                \State $\boldsymbol{C}^k = 
                            \frac{1}{N} \sum_{i}^{N} 
                                \boldsymbol{d}_{Ci}\boldsymbol{d}_{Ci}^T$ \label{last-row}
                \For{$i = 1$ \textbf{to} $N$}
                    \color{red}\State $m = 
                                            \left(
                                                \boldsymbol{\mu}^{k} \odot \boldsymbol{s}
                                            \right)^T 
                                            {\boldsymbol{C}^k}^{-1} 
                                            \left(
                                                \boldsymbol{\mu}^{k} \odot \boldsymbol{s}
                                            \right)$
                    \State $m = m\text{.clamp(min=1)}$
                    \State \color{black}$\alpha_i^{k} = 
                                \max 
                                    \left(
                                        \frac{
                                            \left(
                                                \boldsymbol{L}_i - \boldsymbol{\mu}^k
                                            \right)^T 
                                            {\boldsymbol{C}^k}^{-1} 
                                            \left(
                                                \boldsymbol{\mu}^{k} \odot \boldsymbol{s}
                                            \right) 
                                            - w_i^k
                                        }{
                                            r_i 
                                            \textcolor{red}{m}
                                        }
                                    , 0
                                    \right)$ \label{core-principle}
                \EndFor
            \EndFor
            \State \textbf{return} $\boldsymbol{\alpha}^{N_{\mathrm{iter}}}$
            \EndProcedure
        \end{algorithmic}
    \end{algorithm}
\vspace{-10mm}
\end{figure}
\vspace{-3mm}
\begin{figure}[ht!]
    \vspace{-5mm}
    \centering
    \begin{algorithm}[H]
        \caption{Mag1c-SAS algorithm. The computationally intensive parameter estimation is performed only on a small fraction of the data and then reused in a lightweight filter. Changes to the original Mag1c are highlighted in \textcolor{ForestGreen}{green color}.}
        \label{alg:mag1c-sas}
        \begin{algorithmic}[1]
            \Procedure{Mag1c\textcolor{ForestGreen}{-SAS}}{$\boldsymbol{D}$, $\boldsymbol{s}$, $N_{\mathrm{iter}}$, \textcolor{ForestGreen}{$f=0.01$}}
                \color{ForestGreen}\State Reshape $\boldsymbol{D}$ into $H \cdot W \times p$ matrix
                \State Sample a fraction $f$ of $\boldsymbol{D}$ $\to$ $\boldsymbol{D}_{\text{sample}}$
                 \State $\boldsymbol{\mu}, \boldsymbol{t}, m \gets \textsc{ComputeParameters}(\boldsymbol{D}_{\text{sample}}, \boldsymbol{s}, N_{\mathrm{iter}})$
                \State $\boldsymbol{A} \gets \textsc{LightWeightFilter}(\boldsymbol{D}, N_{\mathrm{iter}}, \boldsymbol{\mu}, \boldsymbol{t}, m)$
                \State Reshape $\boldsymbol{A}$ into $H \times  W$ matrix
                \color{Black}
                \State \textbf{return} $\boldsymbol{A}$
            \EndProcedure
            \State
            \Procedure{\textcolor{ForestGreen}{ComputeParameters}}{$\boldsymbol{L} $, $\boldsymbol{s}$, $N_{\mathrm{iter}}$}
            \State Initialize $\boldsymbol{\mu}^0 = \frac{1}{N} \sum_{i}^{N} \boldsymbol{L}_i$
            \State Initialize $\boldsymbol{C}^0 =
                                \frac{1}{N} \sum_{i}^{N}
                                \left(
                                    \boldsymbol{L}_i - \boldsymbol{\mu}^0
                                \right)
                                \left(
                                    \boldsymbol{L}_i   - \boldsymbol{\mu}^0
                                \right)^T$
            \vspace{1mm}
            \For{$i = 1$ \textbf{to} $N$}
                \State Set $r_i = \frac{
                                        \boldsymbol{L}_i^T \boldsymbol{\mu}
                                    }{
                                        \boldsymbol{\mu}^T \boldsymbol{\mu}
                                    }$
                \State Initialize $ \alpha^0_i = 
                                        \frac{
                                            \left(
                                                \boldsymbol{L}_i - \boldsymbol{\mu}^0
                                            \right)^T 
                                            {\boldsymbol{C}^0}^{-1} 
                                            \left(
                                                \boldsymbol{\mu}^0 \odot \boldsymbol{s} 
                                            \right)
                                        }{
                                            r_i
                                            \left(
                                                \boldsymbol{\mu}^0 \odot \boldsymbol{s}
                                            \right)^T 
                                            {\boldsymbol{C}^0}^{-1} 
                                            \left(
                                                \boldsymbol{\mu}^0 \odot \boldsymbol{s}
                                            \right)
                                        } $ 
            \EndFor
            \For{$k = 1$ \textbf{to} $N_{\mathrm{iter}}$} 
                \State $\boldsymbol{w}^k = \frac{1}{r (\boldsymbol{\alpha}^{k-1} + \epsilon)}$
                \State $\boldsymbol{\mu}^k = 
                            \frac{1}{N} \sum_{i}^{N} 
                                \left(
                                    \boldsymbol{L}_i - r_i \alpha^{k-1}_i \boldsymbol{\mu}^{k-1} \odot \boldsymbol{s}   
                                \right)$
                \For{$i = 1$ \textbf{to} $N$}
                    \State Let $\boldsymbol{d}_{Ci} = 
                        \boldsymbol{L}_i 
                        - r_i \alpha^{k-1}_i \boldsymbol{\mu}^{k} \odot \boldsymbol{s}  
                        - \boldsymbol{\mu}^k$
                \EndFor
                \State $\boldsymbol{C}^k = 
                            \frac{1}{N} \sum_{i}^{N} 
                                \boldsymbol{d}_{Ci}\boldsymbol{d}_{Ci}^T$
                \For{$i = 1$ \textbf{to} $N$}
                    \State $m = 
                                            \left(
                                                \boldsymbol{\mu}^{k} \odot \boldsymbol{s}
                                            \right)^T 
                                            {\boldsymbol{C}^k}^{-1} 
                                            \left(
                                                \boldsymbol{\mu}^{k} \odot \boldsymbol{s}
                                            \right)$
                    \State $m = m\text{.clamp(min=1)}$
                    \State $\alpha_i^{k} = 
                                \max 
                                    \left(
                                        \frac{
                                            \left(
                                                \boldsymbol{L}_i - \boldsymbol{\mu}^k
                                            \right)^T 
                                            {\boldsymbol{C}^k}^{-1} 
                                            \left(
                                                \boldsymbol{\mu}^{k} \odot \boldsymbol{s}
                                            \right) 
                                            - w_i^k
                                        }{
                                            r_i 
                                            m
                                        }
                                    , 0
                                    \right)$
                \EndFor
            \EndFor
            \color{ForestGreen}
            \State $\boldsymbol{t} = {\boldsymbol{C}^{N_{\mathrm{iter}}}}^{-1} 
    \left(
        \boldsymbol{\mu}^{N_{\mathrm{iter}}} \odot \boldsymbol{s}
    \right)$
            \State $m = \left(
                    \boldsymbol{\mu}^{N_{\mathrm{iter}}} \odot \boldsymbol{s}
                \right)^T 
                {\boldsymbol{t}}$
            \State \textbf{return} $\boldsymbol{\mu}^{N_{\mathrm{iter}}}$, $\boldsymbol{t}$, $m$
            \color{black}
            \EndProcedure
            
            \color{ForestGreen}\Procedure{LightweightFilter}{$\boldsymbol{L} $, $N_{\mathrm{iter}}$, $\boldsymbol{\mu}$, $\boldsymbol{t}$, $m$}
            \For{$i = 1$ \textbf{to} $H \cdot W$}
                \State $r_i = \frac{
                                        \boldsymbol{L}_i^T \boldsymbol{\mu}
                                    }{
                                        \boldsymbol{\mu}^T \boldsymbol{\mu}
                                    }$ \label{alg:albedostep}
                \State $m = m\text{.clamp(min=1)}$
                \State $ \alpha^0_i = 
                                        \frac{
                                            \left(
                                                \boldsymbol{L}_i - \boldsymbol{\mu}
                                            \right)^T 
                                            {\boldsymbol{t}}}{
                                            r_i m
                                        } $ 
            \EndFor
            \For{$k = 1$ \textbf{to} $N_{\mathrm{iter}}$} 
                \State $\boldsymbol{w}^k = \frac{1}{r ( \boldsymbol{\alpha}^{k-1} + \epsilon)}$
                \State $\alpha_i^{k} = 
                            \max 
                                \left(\alpha_i^{0}-
                                    \frac{w_i^k
                                    }{
                                        r_i m 
                                    }
                                , 0
                                \right)$
            \EndFor
            \State \textbf{return} $\boldsymbol{\alpha}^{N_{\mathrm{iter}}}$
            \EndProcedure
        \end{algorithmic}
    \end{algorithm}
\vspace{-10mm}
\end{figure}
\subsection{Datasets}
\subsubsection{STARCOP}
For training and testing the models, we use the STARCOP dataset created in \cite{ruuvzivcka2023semantic}. This dataset consists of $512\times512$ tiles of hyperspectral remote sensing images taken by the airborne sensor AVIRIS-NG~\cite{thorpe2016mapping_AVIRISNG} and their semantic segmentation labels of methane plumes. We use the so-called ``allbands'' variant of the dataset released by \cite{Ruzicka2025HyperspectralViTs}, which has 125 spectral bands (RGB + 122 channels in the methane-relevant range of 1573--2480\,nm). For training, the dataset of 3425 tiles is used; for evaluation, we rely on the STARCOP test set of 342 tiles (these are further stratified into weak and strong plume events using the size of the label).
\subsubsection{\emitdatasetabb}
To evaluate generalizability on spaceborne data, we also test on data from the EMIT sensor \cite{green2022_EMIT}, deployed on the International Space Station. This sensor provides 285 spectral bands across 381--2493\,nm, making it well-suited for methane detection. In particular, we use the non-orthorectified radiance from the L1B data product. Based on such data, we manually created the test dataset containing 52 scenes, half with methane and half without. Most of these (45 scenes) are $1280\times1242$ pixels in size, while the rest are larger, reaching up to $2464\times1242$. Of the scenes without methane, half were chosen randomly, and the other half were specifically selected from areas with known methane confounders like cities or solar panel farms. Methane-positive pixels constitute approximately 0.46\,\% of pixels in methane-containing scenes (0.99\,\% for strong leaks and 0.14\,\% for weak leaks). 

\begin{figure}[h]
\centerline{\includegraphics[width=0.5\textwidth]{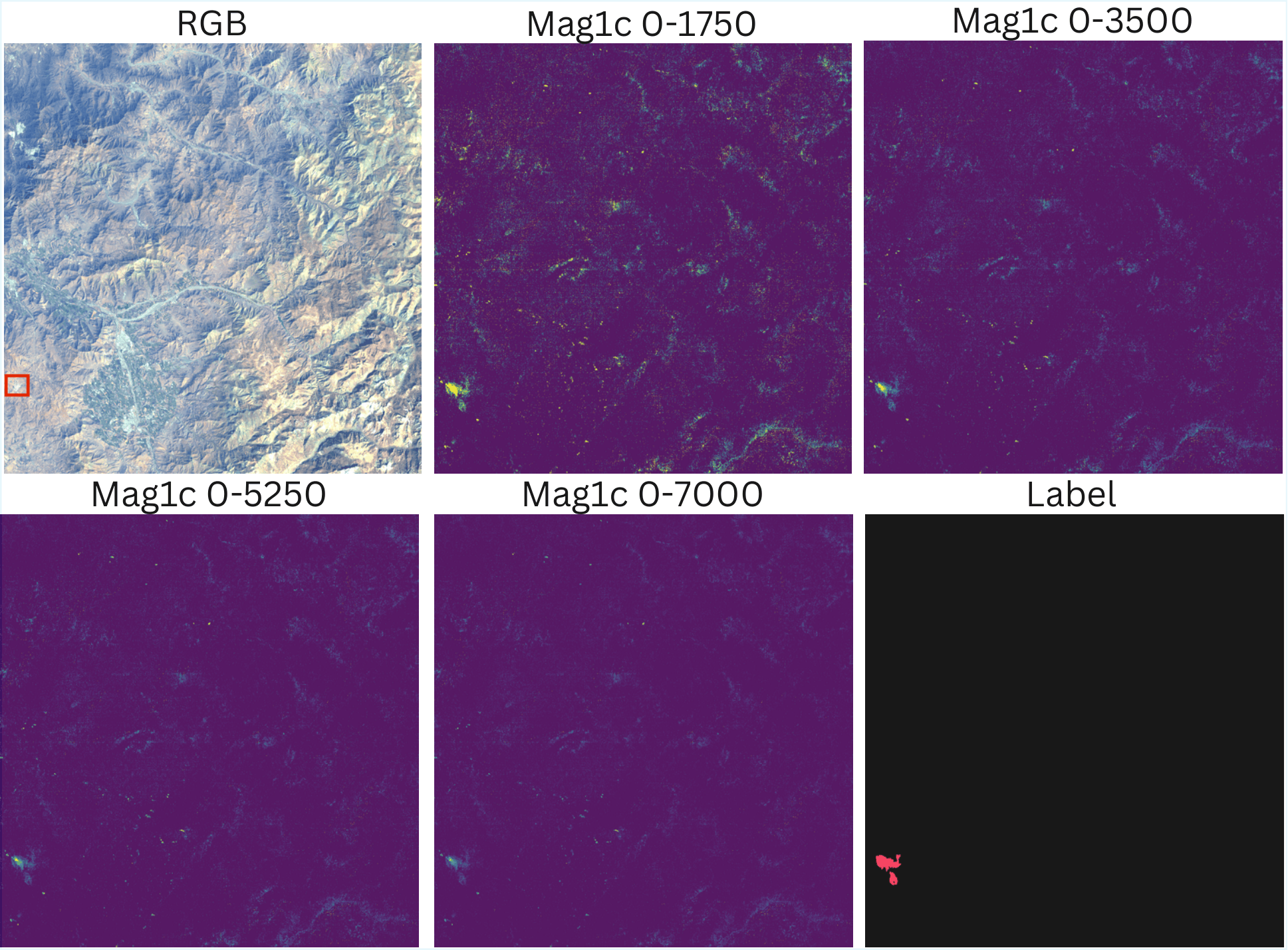}}
\caption{The annotation set uses multiple visualization ranges: lower maximum concentration values reveal all potential positives, while higher values help distinguish true from false positives. The methane leak source area is marked via a red bounding box in the RGB representation.}
\label{annotation-set}
\vspace{-4mm}
\end{figure}

\begin{figure}[h]
\centerline{\includegraphics[width=0.5\textwidth]{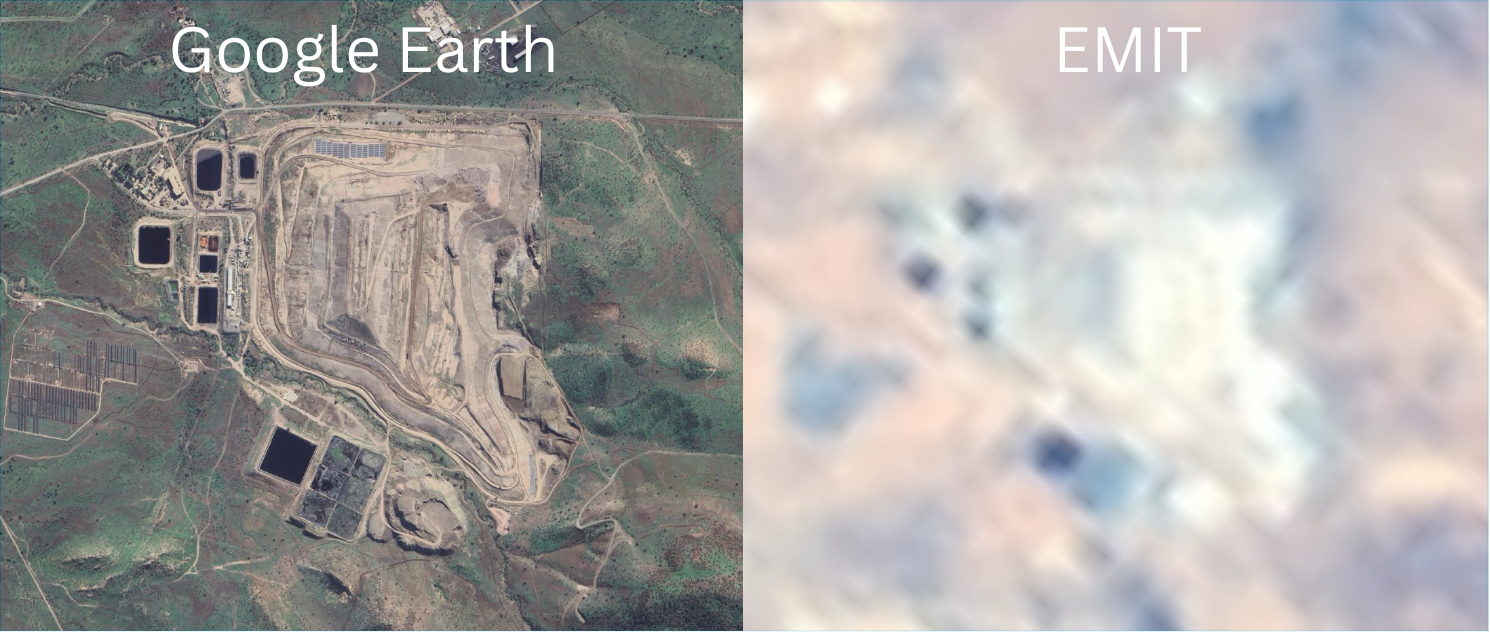}}
\caption{EMIT scene cut-out comparison with high-res imagery from Google Earth. This is the methane leak source area from Figure \ref{annotation-set}. In this case, it helped to confirm that the Mag1c detections are true methane leaks from a landfill.}
\label{google}
\vspace{-4mm}
\end{figure}

We also provide \emph{methane segmentation labels}, which were manually annotated using the Computer Vision Annotation Tool (CVAT)~\cite{boris_sekachev_2020_4009388}. During annotation, RGB imagery and Mag1c visualizations linearly scaled to a $[0, 1]$ range using four different maximum concentration values were uploaded into a single CVAT task for each scene, as shown in Figure \ref{annotation-set}. This setup allows the annotator to switch between views while zooming and navigating within the scene: lower maximum concentration values reveal all candidate methane detections, while higher values help distinguish true methane enhancements from false positives. Additionally, a KML file was provided so the annotator could inspect scene locations in Google Earth when needed. Due to EMIT's 60\,m ground sampling distance, facilities such as mines or gas processing plants may appear only as small traces in the imagery, while high-resolution imagery clearly reveals the infrastructure, helping to distinguish between false and true positive methane leaks, as shown in Figure \ref{google}. After the final scene was annotated, all annotations were reviewed and refined by the annotator to ensure consistency across the dataset, as early annotations may have been of lower quality due to the natural learning curve. The completed annotations were subsequently verified by two additional remote sensing experts.
\vspace{-3mm}
\vspace{-1mm}

\subsection{Band selection}
\begin{figure}[H]
\centerline{\includegraphics[width=0.5\textwidth]{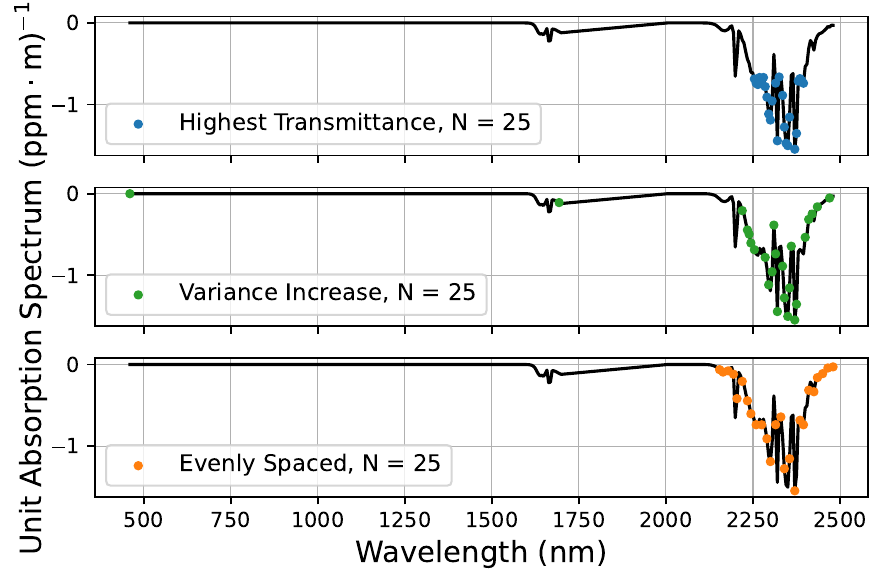}}
\caption{Comparison of selection strategies for the STARCOP data target spectrum.}
\label{selection-strategy}
\vspace{-3mm}
\end{figure}

A crucial hyperparameter is the number of bands used and the method used for their sampling. Generally, using more bands can result in better accuracy but also longer processing time. Therefore, finding the optimal strategy and number of bands is essential. We test three different strategies, visualized in Figure~\ref{selection-strategy}, which involve selecting $N$ bands based on:
\begin{itemize}
\item \textbf{Highest (absolute) transmittance:} Select the bands with the highest absolute transmittance, as these bands have the strongest methane signal.
\item \textbf{Variance increase:} Start with the band that has the highest transmittance, and select each subsequent band to maximize the variance in CH$_4$ transmittance. This allows for a more accurate approximation of the methane transmittance function.
\item \textbf{Evenly spaced in the main methane transmittance range:} Select bands in \tildecorrect2122--2488\,nm range with even spacing between them. \textit{This can be scaled up to 72 bands, at which point all AVIRIS-NG bands within this range are used, as in the original Mag1c.}

\end{itemize}

\subsection{Inference}\label{inference} 
During inference, we compute the methane enhancement product using $N$ selected bands. 
Then, the following two inference strategies (Morphological baseline and ML model) are used:

\subsubsection*{Morphological baseline} The product is first thresholded using a threshold independently optimized for each method on the respective test sets, ensuring a fair comparison. Then, it is processed with morphological opening (erosion followed by dilation), which reduces salt-and-pepper noise.
This approach has been used as a classical non-ML baseline in prior literature~\cite{ruuvzivcka2023semantic}.

\subsubsection*{ML model}
Inspired by \cite{ruuvzivcka2023semantic}, the inputs consist of RGB bands along with the methane enhancement product. The idea is that these products may contain noise or false positives (e.g., solar panels), which can be identified in the RGB bands. The training process is practically the same as in \cite{ruuvzivcka2023semantic}, except that different methane products are used. Namely, the training dataset is tiled into $128\times128$\,px patches with an overlap of 64\,px.
These tiles are sampled with the PyTorch WeightedRandomSampler to balance the number of tiles with and without methane leak events. For data augmentation, we use random rotations, horizontal and vertical flips. Finally, during training, we use the binary cross-entropy loss multiplied by the Mag1c product provided by \cite{ruuvzivcka2023semantic}. We highlight that this is used only during training to guide the model towards regions with strong methane plume events and also confounder areas with strong false signals in the computed methane products. All models were implemented via the Segmentation Models Pytorch library~\cite{SMP}, which allows for easy combination of state-of-the-art encoders and decoders. Namely, we use:
    \begin{itemize}
    \item \textbf{U-Net:}
    As a primary model, we use a classic U-Net architecture \cite{ronneberger2015u} with the fast MobileNet-v2 encoder \cite{sandler2018mobilenetv2}, resulting in 6.6 million parameters and a 25.27\,MB size.
    \item \textbf{LinkNet:}
    As the secondary model, we use LinkNet \cite{chaurasia2017linknet}. It has an Encoder-Decoder-like architecture with skip connections similar to U-Net, but it is designed to be smaller and faster, mainly by using element-wise addition in skip connections instead of concatenation. We also use a very small version of MobileNetv3\cite{mobilenet3} encoder -- \texttt{timm-mobilenetv3\_small\_minimal\_100}, designed in the PyTorch Image Models (timm) library~\cite{timm}. This encoder-decoder combination resulted in only 0.851 million parameters and 3.34\,MB size.
    \end{itemize}
During inference, the model outputs per-pixel probabilities and pixels are classified as methane-positive at a threshold of $p = 0.5$, consistent with the training objective.

\subsection{Computational environment}\label{computational-environment}
The models are trained on a high-performance computing (HPC) cluster with NVIDIA Tesla V100 GPUs. Typical training takes around 8 hours (depending on small IO fluctuations). 

The runtime is measured on the space-qualified Xiphos Q8J board with a flight heritage of over 100 units in orbit, equipped with 4\,GB of RAM and a 4-core Cortex-A53 CPU operating at 1.2\,GHz~\cite{xiphos_q8}. By testing directly on this architecture, we provide a highly accurate evaluation of the computational power actively available on modern satellite missions, such as Phi-Sat-1 \cite{giuffrida2021phi_sat_1_mission} and D-Orbit's ION-SCV satellites \cite{TrainOnBoard}.

Execution times may be marginally affected by differences in library implementations. The baseline algorithms (ACE, CEM, MF) were implemented in NumPy and SciPy, whereas the original Mag1c utilizes PyTorch. To optimize for lightweight deployment, we reimplemented both the tile-wise Mag1c and Mag1c-SAS entirely in NumPy. These NumPy operations were hardware-accelerated using OpenBLAS, which, when compiled without LAPACK, results in a minimal library footprint of only 3.9\,MB. The ML models were compiled using ONNX and deployed using ONNX Runtime~\cite{onnxruntime}.

\section{Results -- Methane segmentation}
\begin{figure*}[p!]
    \centering
    \includegraphics[width=\textwidth]{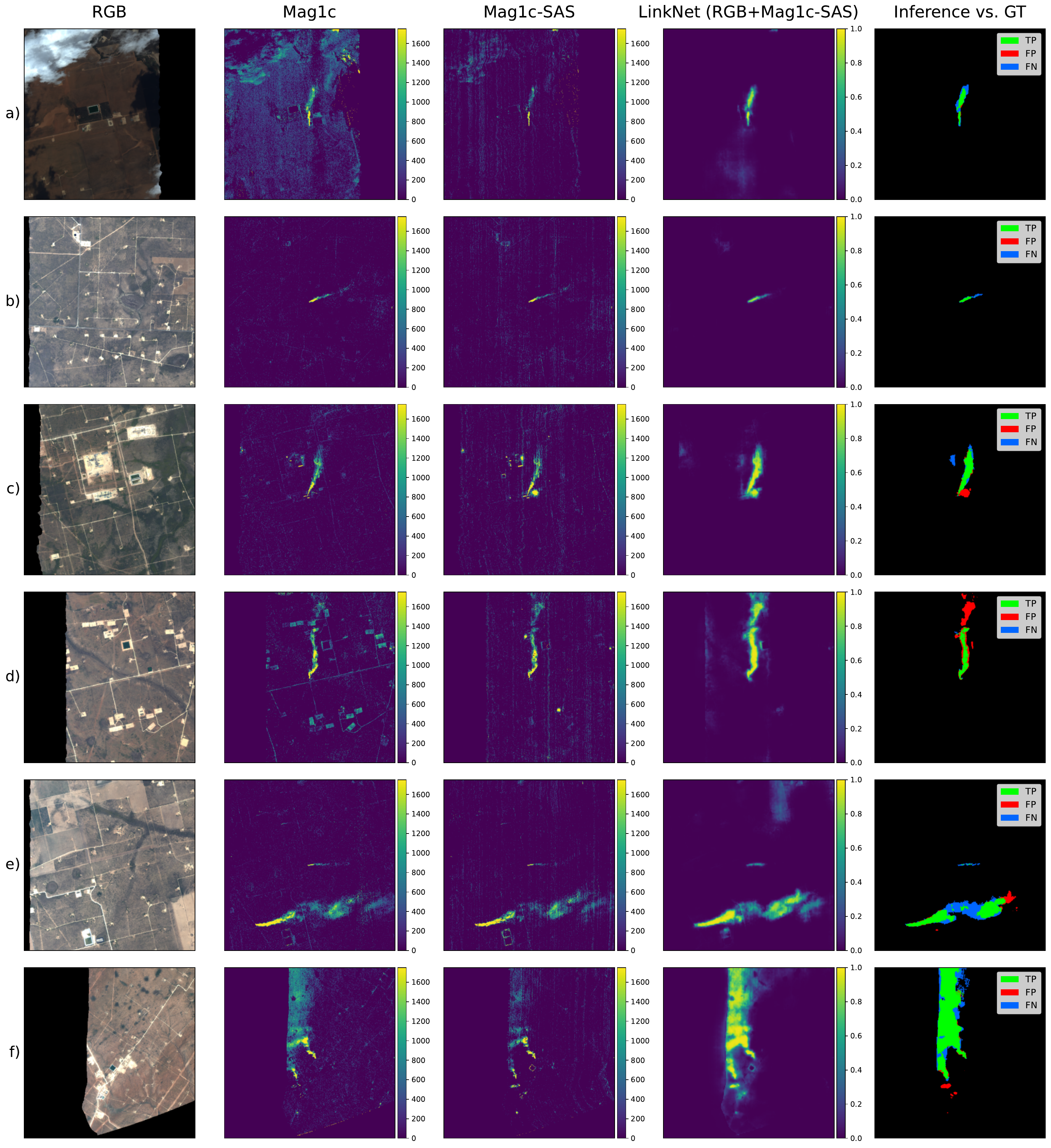}
    \caption{\textbf{Model inference on STARCOP (AVIRIS-NG) data.} Mag1c produces clean methane detections with smooth, plume-like structures. Mag1c-SAS appears more fragmented and noisy, but this is effectively refined by LinkNet. In rows a) and d), Mag1c-SAS significantly reduces false positives, likely due to its larger context window. In other cases, however, it introduces noisy detections -- likely false positives (FP). Some of these are filtered out by the model, as seen in row d), while others remain, such as in row c). In row d), the potential plume tail has limited visibility in the original Mag1c and was not annotated. However, Mag1c-SAS detected it more robustly, resulting in false positives that may reflect true detections.}
    \label{fig:linknet_results}
\end{figure*}
\begin{table*}[ht!]
\caption{Results showing runtimes and metrics for $512\times512$ tiles with 72 bands (entire 2122--2488\,nm range). The total runtime is determined by adding the inference time to the optimized runtime (or, if an optimization was not implemented, the original runtime). For entries that do not mention an ML model, the morphological baseline was applied for inference. The average score and standard deviation are shown for 5 repeated training runs. The runtime is a median of 5 runs. See Section~\ref{computational-environment} for information essential to interpreting the reported runtime.} \label{tab:main-table}
\vspace{-1mm}
\begin{tabularx}{\textwidth}{@{}l*{7}{>{\centering\arraybackslash}X}@{}}
\toprule
\textbf{Method} & \textbf{Recall} & \textbf{Precision} & \textbf{F1} & \textbf{F1 (Strong plumes)} & \textbf{\makecell{Runtime [s]}} & \textbf{\makecell{Inference [s]}}\\ \midrule
Mag1c\,(original, column-wise) & 58.42 & 30.57 & 40.14 & 67.50 & 98.52 & + 0.06 \\ \midrule
Mag1c\,(tile-wise) & 43.71 & 22.79 & 29.96 & 51.77 & 60.91 & + 0.06 \\
U-Net\,[RGB+Mag1c (tile-wise)]& \textbf{65.93 ± 4.04} & 29.51 ± 6.06 & 40.29 ± 5.49 & \textbf{65.34 ± 3.4} & 60.91 & + 2.78 \\ \midrule
Mag1c-SAS& 52.80 & 19.44 & 28.42 & 56.34 & 1.19 & + 0.06 \\
U-Net\,(RGB+Mag1c-SAS)& 56.41 ± 7.0 & 34.62 ± 7.4 & 42.54 ± 6.7 & 61.38 ± 7.7 & 1.19 & + 2.78 \\
ACE & 31.56 & 13.70 & 19.11 & 37.50 & 2.83 & + 0.06 \\
U-Net\,(RGB+ACE) & 47.5 ± 7.6 & 33.72 ± 4.3 & 38.72 ± 1.84 & 51.32 ± 5.09 & 2.83 & + 2.78 \\
CEM & 39.92 & 11.42 & 17.76 & 39.25 & 0.68 & + 0.06 \\
U-Net\,(RGB+CEM) & 55.47 ± 8.4 & 23.49 ± 6.3 & 31.90 ± 4.9 & 53.55 ± 4.9 & 0.68 & + 2.78 \\
MF & 43.73 & 14.07 & 21.29 & 43.96 & 1.11 & + 0.06 \\
U-Net\,(RGB+MF) & 57.41 ± 8.92 & 36.05 ± 8.43 & 43.09 ± 5.68 & 61.32 ± 5.31 & 1.11 & + 2.78 \\
\midrule
LinkNet\,(RGB+Mag1c-SAS) & 51.11 ± 7.2 & \textbf{40.43 ± 6.4} & \textbf{44.44 ± 3.9} & 60.37 ± 5.1 & 1.19 & + 0.33 \\
\midrule
\bottomrule
\end{tabularx}
\vspace{-5mm}
\end{table*}
The results can be seen in Table \ref{tab:main-table}. For strong plumes, the original Mag1c method achieves the highest F1 score (67.50\,\%) but has a very long runtime (98.52\,s). However, even though the runtime is measured here for a $512\times512$ tile for adequate comparison, the product is precomputed in the STARCOP dataset, where it was derived by column-wise application to a whole scene and tiled afterward. Thus, the original Mag1c serves more as an aspirational upper bound rather than a strict baseline.

A more appropriate baseline is the Mag1c applied tile-wise and reimplemented in NumPy, which reduces runtime to 60.91\,s, while also experiencing a significant drop of \tildecorrect16\,pp (percentage points) in F1 score for strong plumes. Interestingly, it was significantly enhanced by the U-Net, which reduces the drop to just \tildecorrect2\,pp.

Among proposed approaches, the arguably strongest one is accelerated Mag1c-SAS, which achieves a drastic \tildecorrect80$\times$ speedup (1.19\,s runtime) while only sacrificing approximately 10\,pp of the strong plume F1 score against the original Mag1c. When combined with the U-Net, this is slightly decreased to just \tildecorrect6\,pp. Interestingly, it has higher training variance (±7.7\,pp) compared to other filter-model combinations, as a probable consequence of the statistical estimation from the sparse subsample.

Among classical target detection methods, ACE is the worst option: it has the longest runtime but doesn't provide the highest performance. In contrast, CEM has the shortest runtime (0.68\,s) but also suffers a significant drop in accuracy. From the simple filters, the best detection performance is demonstrated by MF. When combined with U-Net, the performance is practically the same as for the Mag1c-SAS combination with U-Net. Its runtime is also very close to Mag1c-SAS (1.11\,s vs 1.19\,s), but Mag1c-SAS outperforms MF by \tildecorrect12\,pp in F1 score for strong plumes and by \tildecorrect7\,pp in F1 score for all plumes.
\begin{figure}[h]
\centerline{\includegraphics[width=0.5\textwidth]{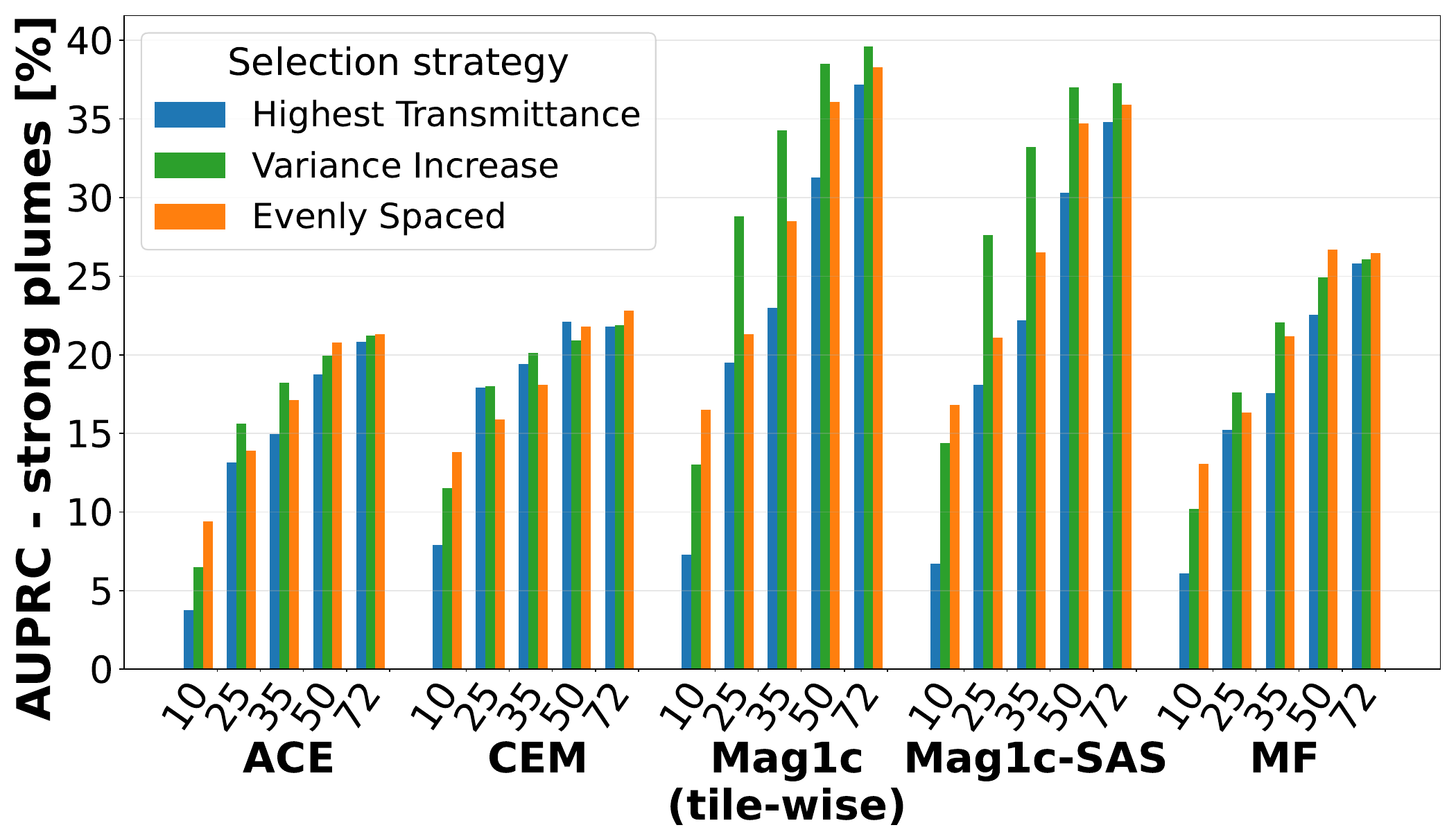}}
\caption{Band selection strategies benchmark.}
\label{select-strategy}
\vspace{-4mm}
\end{figure}

\begin{figure}[h]
\centerline{\includegraphics[width=0.5\textwidth]{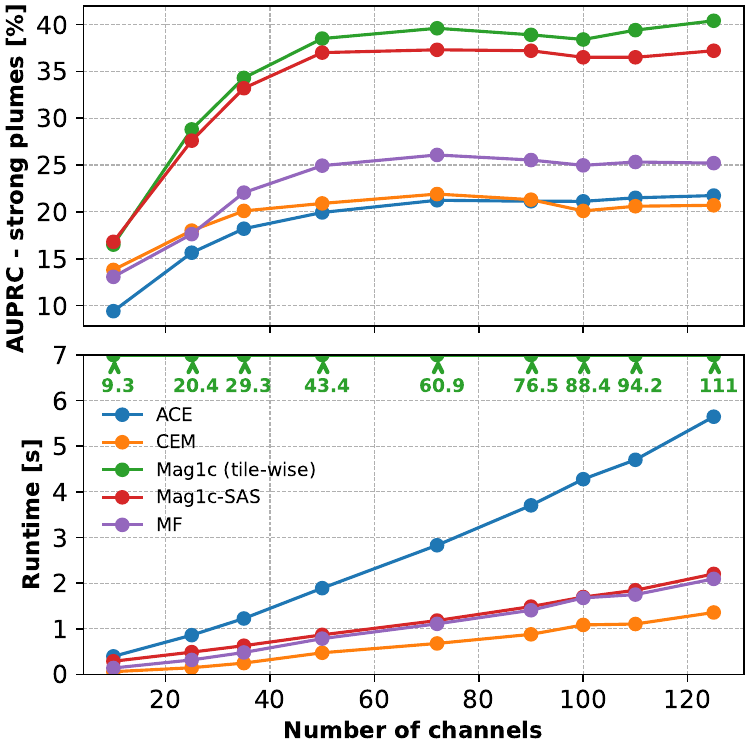}}
\caption{Runtime and AUPRC for strong plumes across different channel numbers. Red crosses indicate the last valid run, as the algorithm ran out of memory for higher channel counts.}
\label{channel-n}
\vspace{-4mm}
\end{figure}

Based on these results, the Mag1c-SAS seems to be the most promising candidate, as it balances accuracy and speed, so we paired it with a more lightweight LinkNet architecture in place of the original U-Net, reducing inference time from 2.78\,s to 0.33\,s. Against the slower U-Net, LinkNet experienced only a minor drop of approximately 1\,pp in F1 score for strong plumes, while the total runtime was reduced from 5.9\,s to 1.58\,s. Interestingly, this configuration received the overall highest F1 score for \textit{all} plumes. Example model predictions can be seen in Figure \ref{fig:linknet_results}.

\section{Results -- Band selection}

All the results in Table \ref{tab:main-table} were computed using 72 spectral channels. To further explore potential runtime reductions, we investigated three band selection strategies, as illustrated in Figure \ref{select-strategy}. For all filters except ACE, the results indicate that for a very small number of channels ($N=10$), the optimal strategy is to select the $N$ channels evenly spaced in the region with strong methane absorption features (\tildecorrect2100--2500\,nm). For a larger number of channels, the best approaches differ; for classical filters, it varies between the Highest Transmittance and Variance Increase, often with negligible difference. Finally, for Mag1c filters, the Variance Increase strategy yields the highest scores.

Also, it seems that using more than 50 channels brings only a negligible improvement, so we examined this for even higher channel counts to verify whether the difference between 50 and 72 wasn’t just a local effect. For this experiment, the Variance Increase selection strategy was used for all channel counts except for $N=10$, where the evenly spaced strategy was employed. The results are in Figure \ref{channel-n}, which suggests that the optimal number of bands for this dataset lies between 50 and 72. Until 50 channels, the score improvements are large; improvements from 50 to 72 channels are only marginal, while beyond 72 channels, the increase is practically non-existent. Meanwhile, runtime continues to rise, indicating diminishing returns in performance relative to computational cost.

\section{Results -- Deployment}
\begin{figure*}[ht!]
    \centering
    \includegraphics[width=\textwidth]{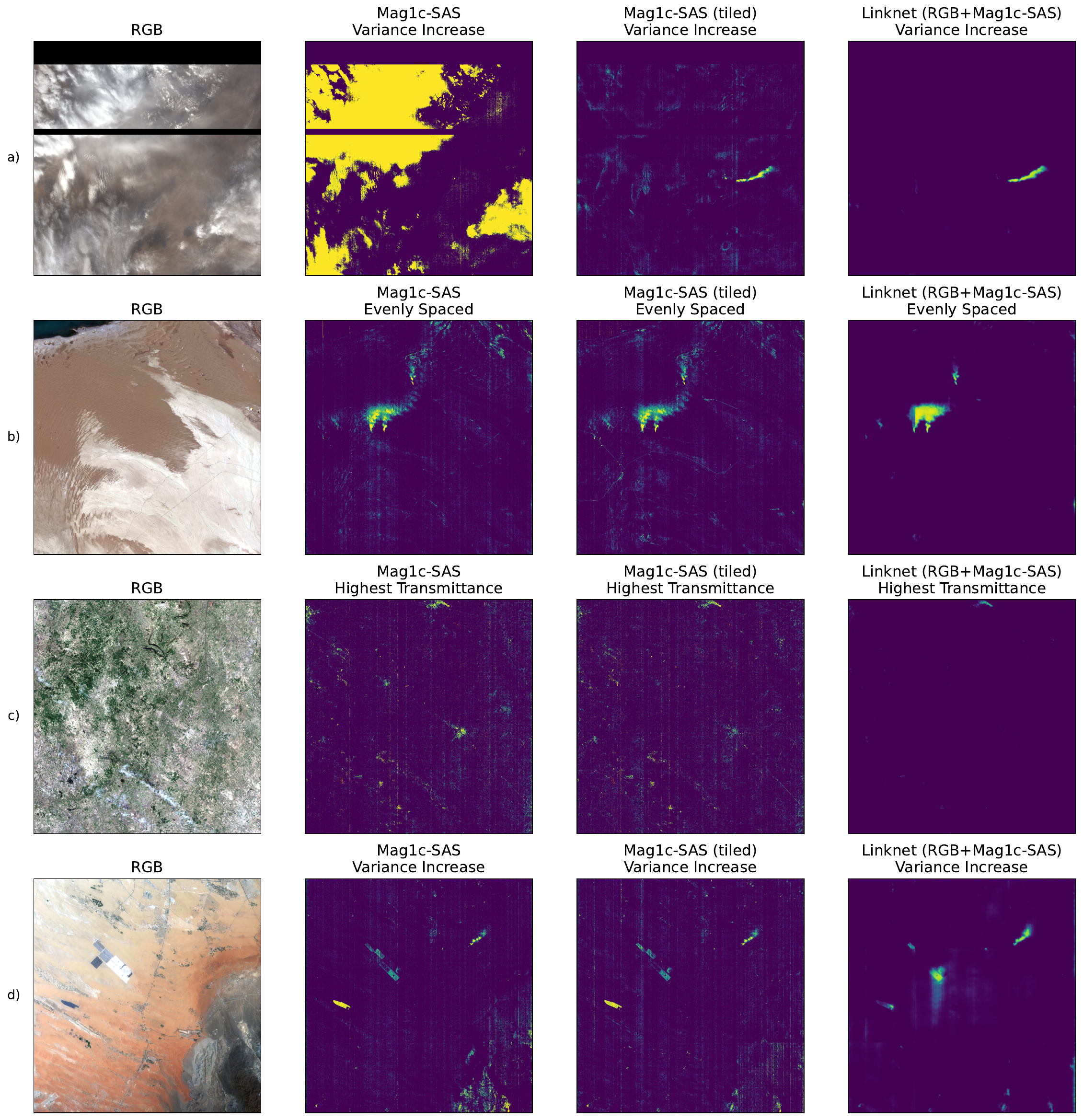}
    \caption{\textbf{Inference on the proposed \emitdatasetabb{} dataset.} (a) Applying Mag1c-SAS to the whole image skews background statistics and produces false positives, which is resolved by processing the image in tiles. (b, c) The LinkNet model filters out noisy Mag1c-SAS detections and identifies both large and small plumes. (d) The model correctly identifies the plume on the right but yields false positives over the solar panels on the left.}
    \label{fig:emit_results}
\end{figure*}
\begin{table*}[ht!]
\centering
\caption{Scores of methods on the \emitdatasetabb{} dataset. For entries that do not mention an ML model, the morphological baseline was applied for inference.}
\label{tab:emit-deployment-metrics}
\begin{tabular}{lllccccc}
\toprule
\textbf{Setup} & \textbf{Method} & \textbf{Band Selection} & \textbf{\shortstack{AUPRC\\{[\%]}}} & \textbf{\shortstack{AUPRC (Strong\\{plumes) [\%]}}} & \textbf{\shortstack{F1\\{[\%]}}} & \textbf{\shortstack{F1 (Strong\\{plumes) [\%]}}} & \textbf{\shortstack{F1 (Classification)\\{[\%]}}} \\
\midrule
-- & Mag1c (original) & Evenly spaced & 6.12 & 35.51 & 47.00 & 62.92 & 69.33 \\
\midrule
\multirow{6}{*}{\shortstack[l]{Mag1c-SAS on tiles\\($512\times512$)}} & \multirow{3}{*}{\shortstack[l]{LinkNet\\(RGB+Mag1c-SAS)}} & Variance Increase & 42.08 & 64.43 & 41.23 & 51.03 & 68.42 \\
& & Highest Transmittance & 41.69 & 64.69 & 38.94 & 49.46 & 70.27 \\
& & Evenly spaced & 40.89 & \textbf{67.47} & 39.52 & 50.89 & 69.33 \\
\cmidrule(lr){2-8}
& \multirow{3}{*}{Mag1c-SAS} & Variance Increase & 3.87 & 28.95 & 29.37 & 58.14 & 66.66 \\
& & Highest Transmittance & 3.36 & 22.74 & 32.69 & 57.34 & 66.66 \\
& & Evenly spaced & 3.34 & 22.24 & 30.63 & 57.08 & 66.66 \\
\midrule
\multirow{6}{*}{\shortstack[l]{Mag1c-SAS\\on whole image}} & \multirow{3}{*}{\shortstack[l]{LinkNet\\(RGB+Mag1c-SAS)}} & Variance Increase & \textbf{42.39} & 53.66 & \textbf{43.56} & 49.62 & 71.23 \\
& & Highest Transmittance & 42.04 & 52.44 & 43.37 & 49.15 & \textbf{73.23} \\
& & Evenly spaced & 41.88 & 54.47 & 42.78 & 50.50 & 68.42 \\
\cmidrule(lr){2-8}
& \multirow{3}{*}{Mag1c-SAS} & Variance Increase & 2.21 & 9.46 & 9.20 & 25.06 & 67.53 \\
& & Highest Transmittance & 5.24 & 33.25 & 36.60 & 60.26 & 66.66 \\
& & Evenly spaced & 2.49 & 31.76 & 11.84 & \textbf{61.05} & 66.66 \\
\bottomrule
\end{tabular}
\end{table*}

\begin{figure}[h]
\centerline{\includegraphics[width=0.5\textwidth]{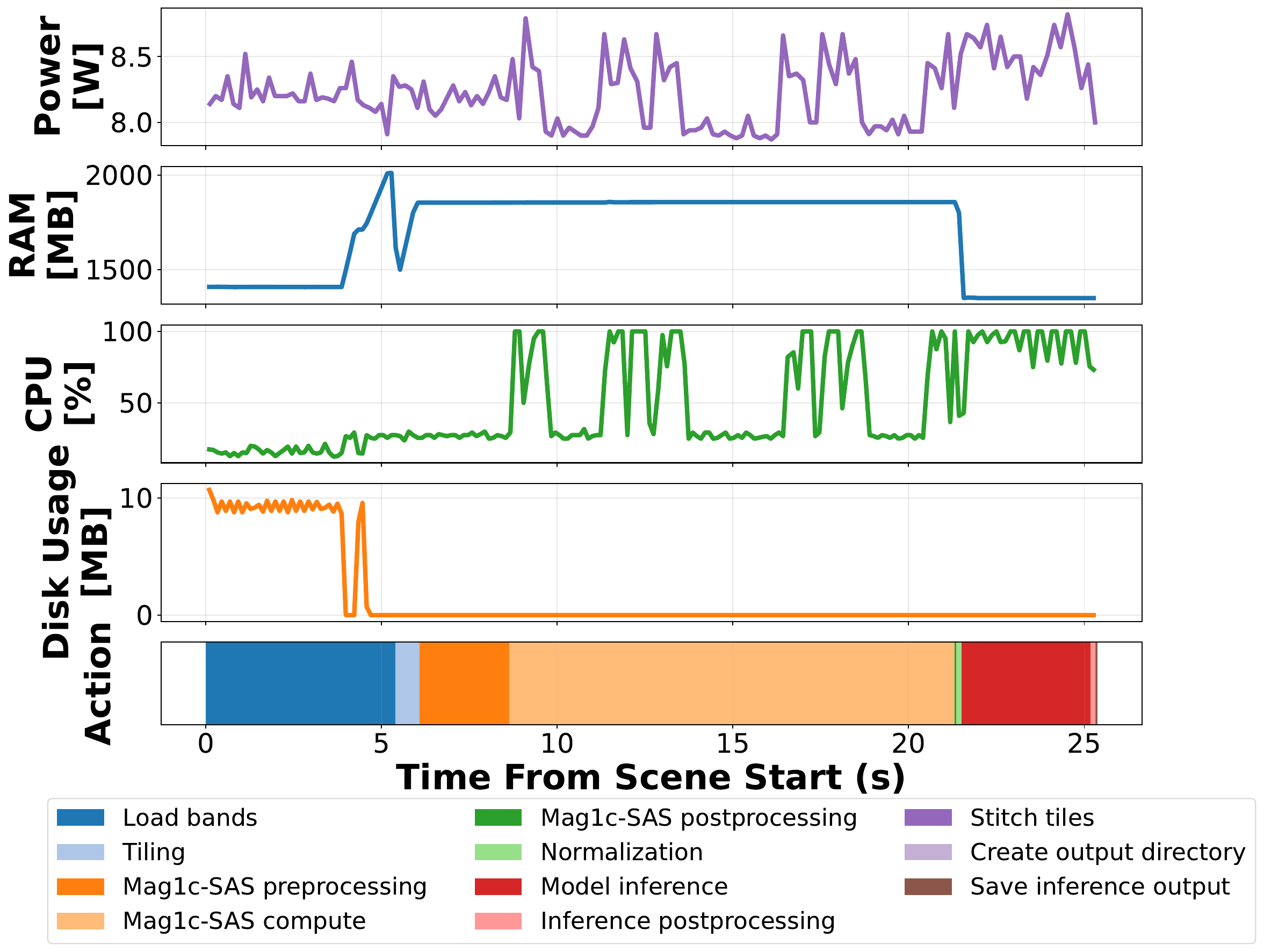}}
\caption{Use of resources by the deployment script with Mag1c-SAS and inference performed both on tiles, shown for a scene with a median runtime.}
\label{benchmarking-telemetry}
\vspace{-4mm}
\end{figure}

\begin{figure}[h]
\centerline{\includegraphics[width=0.5\textwidth]{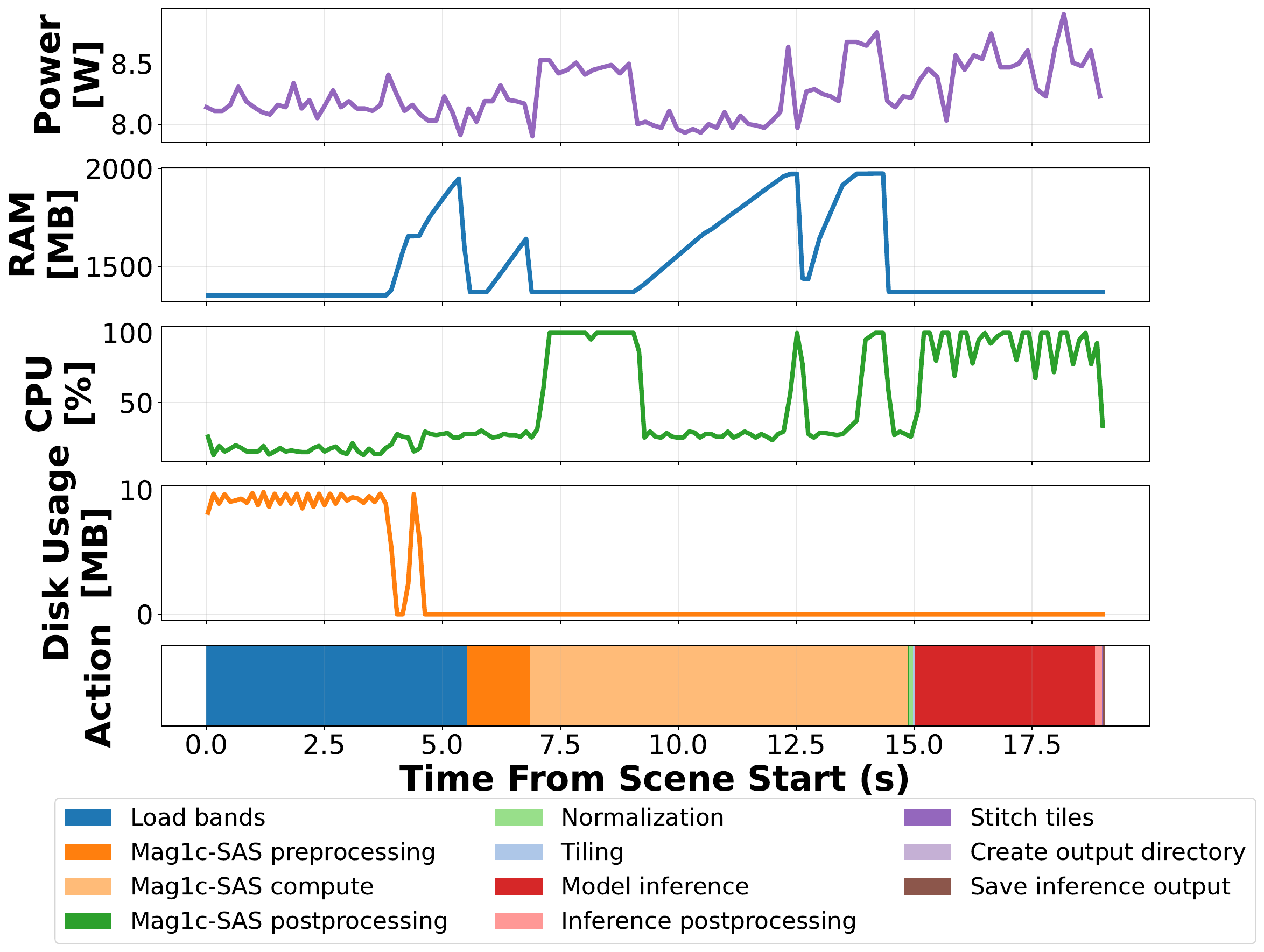}}
\caption{Use of resources by the deployment script with only inference performed on tiles, shown for a scene with a median runtime.}
\label{benchmarking-telemetry_inference_only}
\vspace{-4mm}
\end{figure}

To simulate the deployment, we have created a whole processing pipeline, which was run on all images from the \emitdatasetabb{} dataset. This pipeline begins with loading RGB and methane detection bands, includes all preprocessing and postprocessing steps such as normalization, tiling, and tile stitching, and ends with saving the model inference to the disk. This inference can then be used for ordering or filtering the data for downlink under constrained bandwidth speeds. We have run two variations:
1) Mag1c-SAS and model inference are both done on $512\times512$ tiles;
2) Only the model inference is done on tiles, and Mag1c-SAS is performed on the whole image.
Based on the previous results, namely the segmentation F1 scores in Table \ref{tab:main-table} and AUPRC scores in Figure \ref{select-strategy}, we selected Mag1c-SAS as the filter and the best-performing LinkNet architecture as the model, for their balance of accuracy and speed. Furthermore, based on band selection results, we restricted the number of methane detection bands to 50 and intended to use the Variance Increase selection strategy to select them. However, the \emitdatasetabb{} dataset proved to be different from the AVIRIS-NG STARCOP dataset, as it contains geographically diverse locations, clouds, and relies on a different sensor. Because of this, the Variance Increase strategy, which selects bands from a broader wavelength range, proved to be unstable; consequently, we expanded our evaluation to include the other band selection strategies as well.

The results, detailed in Table \ref{tab:emit-deployment-metrics}, demonstrate that Mag1c-SAS is a highly effective approach. In the best-case scenario (whole image with the Evenly spaced band selection), it yields an F1 (Strong plumes) score of 61.05\,\%, which closely trails the original Mag1c baseline of 62.92\,\%. Interestingly, the standalone Mag1c-SAS can even surpass its LinkNet-paired counterpart in the F1 metric (e.g., 61.05\,\% vs. 50.50\,\% in the Evenly spaced whole-image setup). However, this discrepancy is primarily an artifact of threshold selection. For both Mag1c and Mag1c-SAS, we applied the optimal threshold for the \emitdatasetabb{} dataset, empirically determined to be 1200 (compared to 300 for the STARCOP dataset). Conversely, the LinkNet model probabilities were kept at a 0.5 threshold, with which the model was trained on STARCOP. When the LinkNet threshold is adjusted to 0.16, its F1 (Strong plumes) score increases to a best-case 62.62\,\% (using the Evenly spaced band selection strategy with tiling for both Mag1c-SAS and inference).

When evaluating the standard F1 metric (for all plumes), Mag1c-SAS falls significantly behind LinkNet, particularly when applied to the whole image. For instance, using the Evenly spaced strategy on the whole image, Mag1c-SAS achieves an F1 score of only 11.84\,\%, compared to LinkNet's 42.78\,\%. However, due to the mentioned threshold dependencies, a more robust performance comparison between LinkNet and Mag1c-SAS should rely on the threshold-independent Area Under the Precision-Recall Curve (AUPRC) metrics. Examining these values reveals that the LinkNet model effectively filters out false positives from the Mag1c-SAS outputs. For example, when using the Evenly spaced strategy on tiles, LinkNet achieves the highest overall AUPRC (Strong plumes) of 67.47\,\%, a \tildecorrect45\,pp improvement over the 22.24\,\% achieved by Mag1c-SAS alone.

Furthermore, applying Mag1c-SAS across tiles yields much more stable results than applying it to whole images. The instability in whole images is particularly severe with the Variance Increase band selection strategy, where Mag1c-SAS suffers drastic drops across all segmentation metrics (e.g., regular AUPRC plunges to 2.21\,\% and F1 to 9.20\,\%). The Evenly spaced strategy on whole images exhibits split behavior: it experiences a sharp drop in performance for all plumes (F1 dropping to 11.84\,\%), but simultaneously sees a slight increase for strong plumes, with AUPRC (Strong plumes) rising to 31.76\,\% and F1 (Strong plumes) to 61.05\,\%. Meanwhile, the Highest Transmittance strategy on whole images shows minor increases across all segmentation metrics compared to its tiled counterpart. Ultimately, the optimal AUPRC and AUPRC (Strong plumes) scores are reached by deploying LinkNet on the Mag1c-SAS outputs generated from tiled images. Visualizations of these key results are presented in Figure \ref{fig:emit_results}.

To evaluate runtime and utilization of resources, we tested resource utilization on the Xiphos Q8J using both a full tiling configuration (applied to both Mag1c-SAS and inference) and an inference-only tiling approach. The use of resources is visualized for a representative scene in Figure~\ref{benchmarking-telemetry} and Figure~\ref{benchmarking-telemetry_inference_only}. Regarding processing time, the full tiling approach for the entire dataset of 52 scenes (totaling 103.75\,GB across all EMIT 285 bands) required 1407.89\,seconds, yielding an average processing time of 27.07 seconds per scene and a throughput of 73.69\,MB/s. Restricting tiling strictly to the inference stage significantly improved overall speed, reducing the total processing time to 1069.20\,seconds, or 20.55\,seconds per scene on average, which effectively boosted the throughput to 97.03\,MB/s. A breakdown of the execution pipeline indicates that Mag1c-SAS computation remains the most time-intensive stage in both scenarios; however, omitting tiling from this stage reduced its average computational time from 12.69\,seconds to 8.49\,seconds per scene. Other structural overheads remained highly consistent across both setups, with tile-based model inference consuming roughly 4.02 to 4.07\,seconds and initial scene loading taking approximately 6.16 to 6.24\,seconds per scene. Power consumption and computational demands exhibited minimal variance between the two methods. Before execution, the system consumed an average of 7.71\,W in idle mode, while during the active running phase, average power consumption increased slightly to 8.20\,W (maximum 9.14\,W) for full tiling and 8.24\,W (maximum 9.07\,W) for inference-only tiling. Computational demands were moderate in both setups, with average CPU utilization ranging from 47.5\,\% to 48.1\,\%, accompanied by temporary spikes to 100\,\% during computationally demanding phases. Memory (RAM) usage showed a slight trade-off. Full tiling maintained a higher average of 1874.74\,MB but had a lower peak of 2891.29\,MB. Meanwhile, inference-only tiling operated at a lower average of 1665.43\,MB but hit a higher peak of 3387.99\,MB. This happens because most scenes in the dataset are 1242 pixels tall, which mostly drives the average, while the few larger scenes are responsible for those peaks. Finally, disk usage proved exceptionally lightweight across the board, averaging between 1.58\,MB and 2.14\,MB for read operations and remaining negligible (0.03\,MB to 0.04\,MB) for write operations.

\section{Discussion}

Our research tested and sped up various filters for onboard methane detection. The Mag1c-SAS filter was the most promising because it is fast enough and produces results visually similar to the original Mag1c approach, though slightly noisier. We fixed this drawback by pairing it with a lightweight LinkNet model to clean the outputs, keeping the final detections both fast and accurate. On the two datasets we tested, Mag1c-SAS combined with LinkNet delivered results that were comparable to, and sometimes even better than, slower Mag1c. Notably, the models were trained exclusively on airborne AVIRIS-NG data, yet still generalized well to the unseen spaceborne EMIT sensor, as demonstrated by high AUPRC scores. However, it was also demonstrated that the segmentation threshold requires tuning when transferring to a different sensor.

We also tested the number of channels and the strategies used to select them. On the STARCOP dataset, we found that using more bands does not necessarily improve the results. Beyond 50 bands, only the processing time continued to rise. The results for band selection strategies were mixed. For classic, generally less accurate methods like MF or ACE, the choice of strategy had minimal impact. However, with the more precise Mag1c algorithms, band selection made a clear difference. Selecting bands that maximized the variance in transmittance proved the most effective. This is likely because greater variance allows for a better approximation of the overall methane transmittance function. However, this was only consistent on the geographically uniform STARCOP dataset, which contains almost no clouds. On the \emitdatasetabb{} dataset, the performance of these selection methods was less clear. When Mag1c-SAS was run on the entire scene rather than smaller tiles, the Variance Increase strategy resulted in imprecise background statistics. This caused false-positive cloud detections and made it the worst-performing method. This was probably because it included bands across a broader wavelength range. The Evenly Spaced strategy, which focuses on the main methane range \tildecorrect2100--2500\,nm, yielded better results, while the Highest Transmittance strategy performed the best overall. It seems that in ideal conditions, the Variance Increase strategy is the top performer, but using bands with the strongest methane absorption is more reliable.

This instability can be mitigated by running Mag1c-SAS on smaller tiles. With smaller tiles, the differences between band selection strategies were negligible, and the AUPRC scores when paired with LinkNet were the highest. This is likely because tiling constrains the background to a smaller, more homogeneous area: when Mag1c-SAS is applied to a full scene, the background can be spatially heterogeneous — pixels from one part of the scene may not be statistically representative of another. This is already a problem on its own, but it is further amplified by the fact that Mag1c-SAS estimates background statistics from only a small fraction of uniformly sampled pixels: in a heterogeneous scene, such a sparse sample can easily fail to capture the local background conditions across the entire image, making the sampled fraction an unreliable basis for parameter estimation. While this tile approach is more stable, it is also noticeably slower. A faster yet still robust alternative is to run Mag1c-SAS on whole scenes while using the Highest Transmittance strategy to select the bands.

Mag1c-SAS shows strong performance relative to the original Mag1c. Nonetheless, the absolute performance remains modest, with F1 scores for strong plumes peaking at around 60\,\% when combined with LinkNet and around 70\,\% for classification F1 score. Due to its higher recall and lower precision, the system could alternatively serve not as a classifier but as a methane extraction algorithm, where the Mag1c-SAS product itself, a 2D compression of the hyperspectral cube, would be transmitted to the ground for further analysis. As an alternative to our filter-based approach, future missions could forgo methane enhancement product computation entirely in favor of end-to-end machine learning (ML) models, as exemplified by \cite{Ruzicka2025HyperspectralViTs}. While end-to-end ML methods might extract methane features better from raw hyperspectral data to achieve higher accuracy, our filter-based approach generalizes much better across different sensors without needing to be retrained. This was proven by its strong performance on the \emitdatasetabb{} dataset, which was unseen during training.

To ensure this generalized approach is highly accessible for mission operators, we have released the proposed pipeline as a lightweight, user-friendly PyPI library that relies exclusively on NumPy and, optionally, ONNX Runtime for inference. Hardware profiling confirms the library's efficiency: it effectively leverages onboard CPU and RAM to achieve high throughput with only a marginal increase in power consumption. Consequently, these algorithms are ready for deployment on low-power satellites without requiring extensive pre-launch training or setup. The achieved throughput suggests the system can process scenes between acquisition windows, though precise real-time viability depends on mission-specific data acquisition rates. They are ideally suited to serve as an initial, broad detection system in orbit. Within a few months, as sufficient orbital data is collected from the specific sensor, this initial system could be replaced by the aforementioned end-to-end ML models fine-tuned on that newly acquired data.

While this research represents a significant step, important avenues for future work remain. Specifically, our algorithms were evaluated on preprocessed datasets and may exhibit degraded performance if applied directly to raw, unprocessed sensor data. A primary vulnerability is spatial band misalignment, as the presented methods rely heavily on precise pixel-level alignment across the entire spectral range. Furthermore, the absence of standard radiometric or atmospheric corrections on raw data would likely introduce significant noise into the detection process. Executing these corrections efficiently on board, given the high dimensionality of hyperspectral data, remains a critical computational challenge that must be addressed before fully autonomous, raw-data methane detection can be realized.

Overall, this research presents fast, nearly out-of-the-box algorithms capable of detecting spectral targets on low-power CPUs, available on board satellites. Combined with our insights into band selection, this work provides a crucial step toward a highly generalizable, fast, and computationally efficient hyperspectral segmentation system. While currently optimized for methane, the core approach of pairing a filter with an ML model has great potential for a wide range of other remote sensing applications simply by adjusting the filter and the target spectrum (as demonstrated for non-onboard deployment in \cite{ruzicka2026fully}).

\section*{Acknowledgment}
The work of J.H. and R.P. was supported by the Technology Agency of the Czech Republic (TA CR) under the TREND Programme, project no. FW09020069. A portion of the work by V.R. was carried out at the Jet Propulsion Laboratory, California Institute of Technology, under a contract with the National Aeronautics and Space Administration (80NM0018D0004) -- this research was supported by an appointment to the NASA Postdoctoral Program at the Jet Propulsion Laboratory, administered by Oak Ridge Associated Universities under contract with NASA.

\bibliographystyle{IEEEtran}
\bibliography{bib}

\vspace{4pt}

\begin{IEEEbiography}
[{\includegraphics[width=1in,height=1.25in,clip,keepaspectratio]{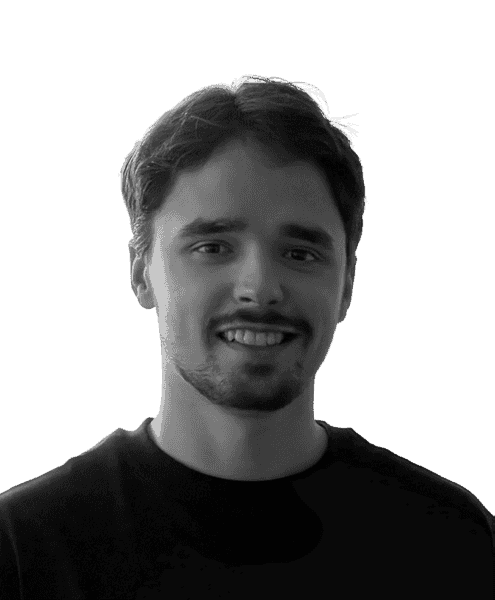}}]{Jonáš Herec}
is a Lead Machine Learning Researcher at Zaitra s.r.o. After earning his graduate degree in Machine Learning and Artificial Intelligence from the Faculty of Informatics at Masaryk University, he is continuing his PhD studies there. The focus of both his work and studies is onboard processing of satellite imagery for time-critical applications.
\end{IEEEbiography}

\vspace{4pt}

\begin{IEEEbiography}
[{\includegraphics[width=1in,height=1.25in,clip,keepaspectratio]{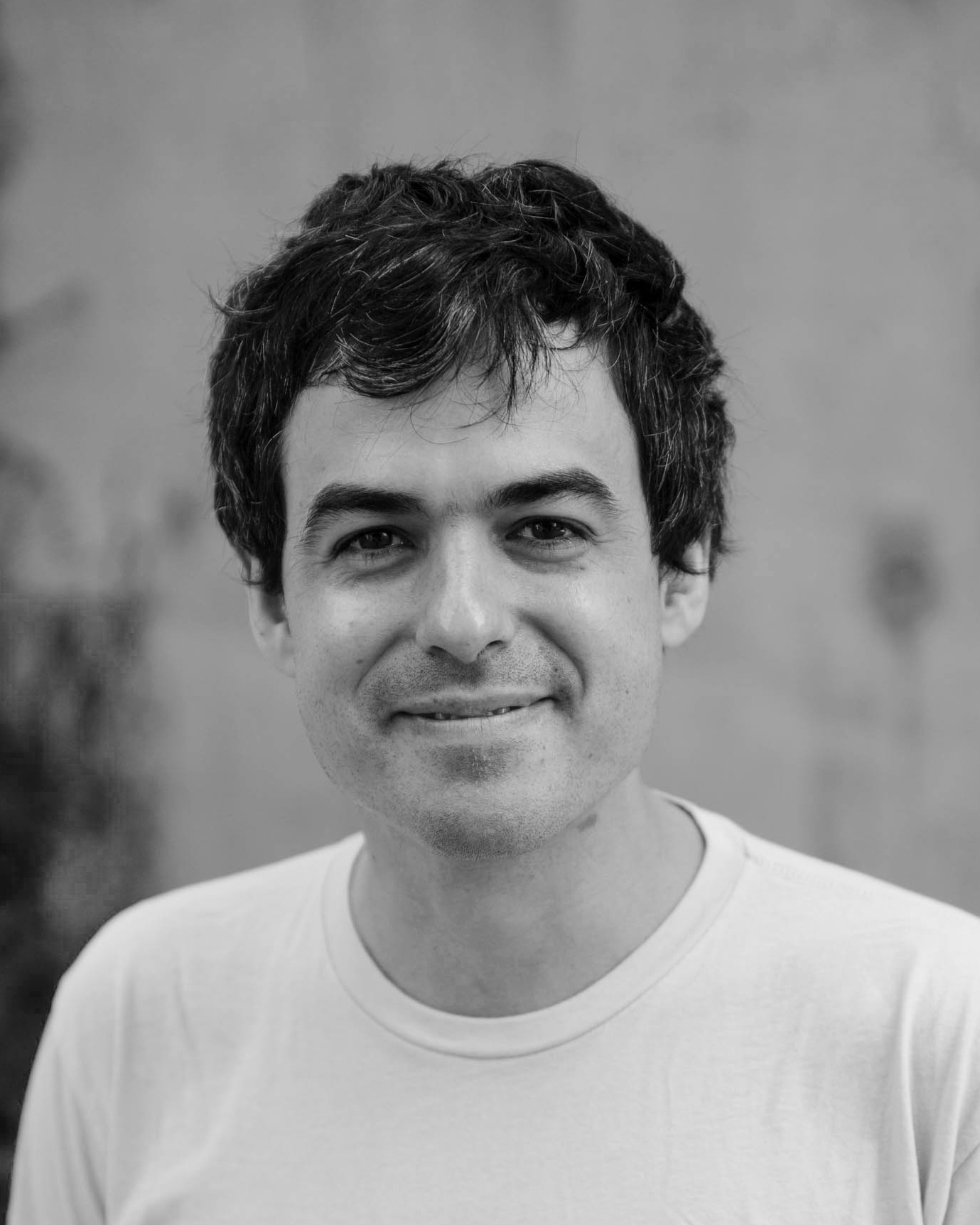}}]{Vít Růžička} received the B.Sc. and M.Sc. degrees in Computer Science from the Czech Technical University, Prague. He received his Ph.D. degree in Computer Science at the University of Oxford, UK, in 2025.
He was a Visiting Researcher with the European Space Agency (2023) and a Consultant for the United Nations Environment Programme (2024).
He is currently a postdoc at NASA Jet Propulsion Lab, supervised by David R. Thompson. His research interests include machine learning for imaging spectroscopy data, in applications connected to climate change and on-board deployment on low compute systems, such as satellites.
\end{IEEEbiography}

\vspace{4pt}

\begin{IEEEbiography}
[{\includegraphics[width=1in,height=1.25in,clip,keepaspectratio]{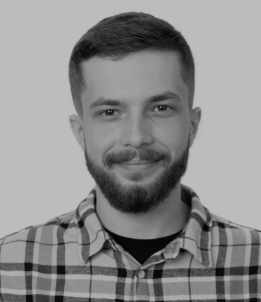}}]{Radoslav Pitoňák}
is the CTO of Zaitra, s.r.o. He holds a Bachelor’s degree in Information Technology, specializing in edge AI and the application of deep learning to satellite imagery processing. He has held engineering roles on ESA projects, including on-board data processing for Earth observation products and studies on autonomous space tugs.
\end{IEEEbiography}

\vspace{4pt}

\begin{IEEEbiography}
[{\includegraphics[width=1in,height=1.25in,clip,keepaspectratio]{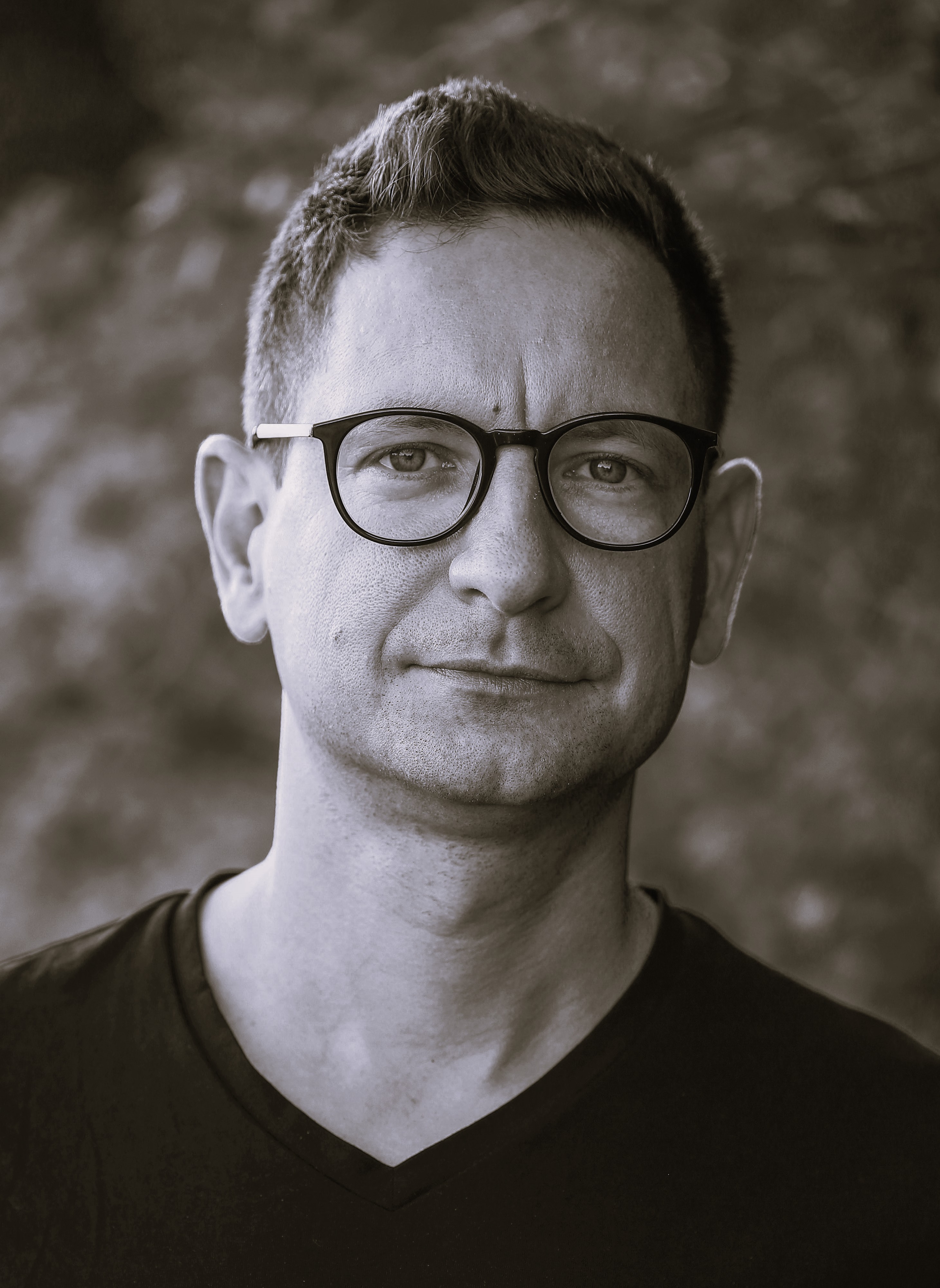}}]{Jan Sedmidubsky}
is an associate professor at the Faculty of Informatics, Masaryk University, where he completed his habilitation in 2021. His research focuses on similarity search, large-scale data analytics, and the processing of multimedia and human-motion data. He has contributed to more than 15 research projects and co-authored more than 50 scientific publications, several of which have received best paper awards (e.g., ISM 2017, SIGIR 2023, SISAP 2024). He also collaborates with industry partners to develop real-world applications.
\end{IEEEbiography}

\vfill

\end{document}